# Part-of-Speech Tagging with Neural Networks

Helmut Schmid

Institute for Computational Linguistics, Azenbergstr.12, 70174 Stuttgart, Germany,
schmid@ims.uni-stuttgart.de

**Topic area:** large text corpora, part-of-speech tagging, neural networks

## 1 Abstract

Text corpora which are tagged with part-of-speech information are useful in many areas of linguistic research. In this paper, a new part-of-speech tagging method based on neural networks (*Net-Tagger*) is presented and its performance is compared to that of a HMM-tagger (Cutting et al., 1992) and a trigram-based tagger (Kempe, 1993). It is shown that the Net-Tagger performs as well as the trigram-based tagger and better than the HMM-tagger.

## 2 Introduction

Words are often ambiguous in their part of speech. The English word *store* for example can be either a noun, a finite verb or an infinitive. In an utterance, this ambiguity is normally resolved by the context of a word: e.g. in the sentence "*The 1977 PCs could store two pages of data.*", *store* can only be an infinitive.

A *part-of-speech tagger* is a system which automatically assigns the part of speech to words using contextual information. Potential applications for part-of-speech taggers exist in many areas including speech recognition, speech synthesis, machine translation and information retrieval.

Different methods have been used for the implementation of part-of-speech taggers. TAGGIT (Greene, Rubin, 1971), an early system, which was used for the initial tagging of the Brown corpus was *rule-based*. It was able to assign the correct part-of-speech to about 77 % of the words in the Brown corpus.

In another approach contextual dependencies are modelled *statistically*. Church (1988) and Kempe (1993) use second order Markov Models and train their systems on large handtagged corpora. Using this method, they are able to tag more than 96 % of their test words with the correct part-of-speech. The need for reliably tagged training data, however, is a problem for languages, where such data is not available in sufficient quantities. Jelinek (1985) and Cutting et al. (1992) circumvent this problem by training their taggers on untagged data using the Baum-Welch algorithm (also know as the forward-backward algorithm). They report rates of correctly tagged words which are comparable to that presented by Church (1988) and Kempe (1993).

A third and rather new approach is tagging with *artificial neural networks*. In the area of speech recognition neural networks have been used for a decade now. They have shown performances comparable to that of Hidden Markov model systems or even better (Lippmann, 1989). Part-of-speech prediction is another area, closer to POS tagging, where neural networks have been applied successfully. Nakamura et al. (1990) trained a 4-layer feed-forward network with up to three preceding part-of-speech tags as input to predict the word category of the next word. The prediction accuracy was similar to that of a trigram-based predictor. Using the predictor, Nakamura et al. were able to improve the recognition rate of their speech recognition system from 81.0 % to 86.9 %.

Federici and Pirrelli (1993) developed a part-of-speech tagger which is based on a special type of neural network. It disambiguates between alternative morphosyntactic tags which are generated by a morphological analyzer. The tagger is trained with an analogy-driven learning procedure. Only preliminary results are presented, so that a comparison with other methods is difficult.

In this paper, a part-of-speech tagger based on a multilayer perceptron network is presented. It is similar to the network of Nakamura et al. (1990) in so far as the same training procedure (Backpropagation) is used; but it differs in the structure of the network and also in its purpose (disambiguation vs. prediction). The performance of the presented tagger is measured and compared to that of two other taggers (Cutting et al., 1992; Kempe, 1993).

## 3 Neural Networks

Artificial neural networks consist of a large number of simple processing units. These units are highly interconnected by directed weighted links. Associated with each unit is an *activation value*. Through the connections, this activation is *propagated* to other units.

In *multilayer perceptron networks* (MLP-networks), the most popular network type, the processing units are arranged vertically in several layers (fig. 1). Connections exist only between units in adjacent layers. The bottom layer is called *input layer*, because the activations of the units in this layer represent the input of the network. Correspondingly, the top layer is called *output layer*. Any layers between input layer

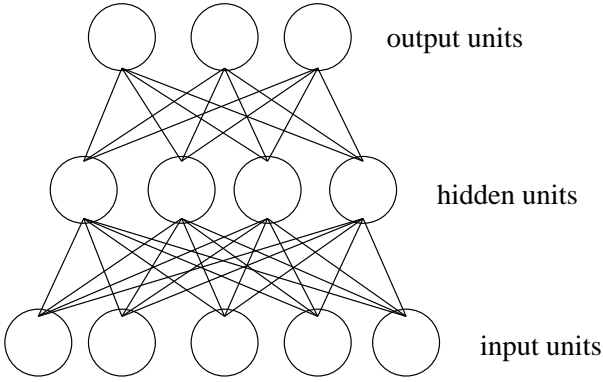

Figure 1: A 3-layer perceptron network

and output layer are called *hidden layers*. Their activations are not visible externally.

During the processing in a MLP-network, activations are propagated from input units through hidden units to output units. At each unit $j$, the weighted input activations $a_i w_{ij}$ are summed and a bias parameter $\theta_j$ is added.

$$net_j = \sum_i a_i w_{ij} + \theta_j \qquad (1)$$

The resulting network input $net_j$ is then passed through a sigmoid function (the *logistic function*) in order to restrict the value range of the resulting activation $a_j$ to the interval [0,1].

$$a_j = \frac{1}{1 + e^{-net_j}} \qquad (2)$$

The network *learns* by adapting the weights of the connections between units, until the correct output is produced. One widely used method is the *backpropagation algorithm* which performs a gradient descent search on the error surface. The weight update $\Delta w_{ij}$, i.e. the difference between the old and the new value of weight $w_{ij}$, is here defined as:

$$\begin{aligned} \Delta w_{ij} &= \eta a_{pi} \delta_{pj}, \quad \text{where} \\ \delta_{pj} &= \begin{cases} a_{pj}(1 - a_{pj})(t_{pj} - a_{pj}), \\ \quad \text{if } j \text{ is an output unit} \\ a_{pj}(1 - a_{pj}) \sum_k \delta_{pk} w_{jk}, \\ \quad \text{if } j \text{ is a hidden unit} \end{cases} \end{aligned} \qquad (3)$$

Here, $t_p$ is the target output vector which the network must learn[1].

Training the MLP-network with the backpropagation rule guarantees that a local minimum of the error surface is found, though this is not necessarily the global one. In order to speed up the training process, a *momentum term* is often introduced into the update formula:

$$\Delta w_{ij}(t+1) = \eta a_{pi} \delta_{pj} + \underline{\alpha \Delta w_{ij}(t)} \qquad (4)$$

---

[1] We assume here that the bias parameter $\theta_j$ is realized as a weight to an additional unit which has always the activation value 1 (cp. (Rumelhart, McClelland, 1984)).

For a detailed introduction to MLP networks see e.g. (Rumelhart, McClelland, 1984).

## 4 THE TAGGER NETWORK

The *Net-Tagger* consists of a MLP-network and a lexicon (see fig. 2).

Figure 2: Structure of the Net-Tagger without hidden layer; the arrow symbolizes the connections between the layers.

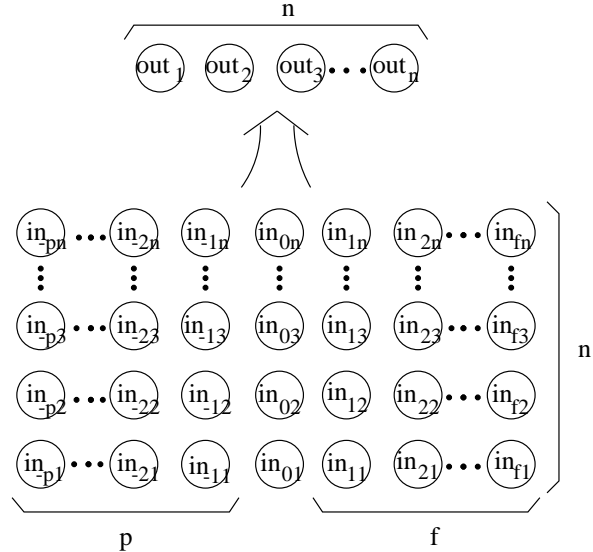

In the output layer of the MLP network, each unit corresponds to one of the tags in the tagset. The network learns during the training to activate that output unit which represents the correct tag and to deactivate all other output units. Hence, in the trained network, the output unit with the highest activation indicates, which tag should be attached to the word that is currently processed.

The input of the network comprises all the information which the system has about the parts of speech of the current word, the $p$ preceding words and the $f$ following words. More precisely, for each part-of-speech tag $pos_j$ and each of the $p+1+f$ words in the context, there is an input unit whose activation $in_{ij}$ represents the probability that $word_i$ has part of speech $pos_j$.

For the word which is being tagged and the following words, the *lexical part-of-speech probability* $P(pos_j | word_i)$ is all we know about the part of speech[2]. This probability does not take into account any contextual influences. So, we get the following input representation for the currently tagged word and the following words:

$$in_{ij} = P(pos_j | word_i), \quad \text{if} \quad i \geq 0 \qquad (5)$$

---

[2] Lexical probabilities are estimated by dividing the number of times a word occurs with a given tag by the overall number of times the word occurs. This method is known as the *Maximum Likelihood Principle*.

For the preceding words, there is more information available, because they have already been tagged. The activation values of the output units at the time of processing are here used instead of the lexical part-of-speech probabilities[3]:

$$in_{ij}(t) = out_j(t+i), \quad \text{if} \quad i < 0 \qquad (6)$$

Copying output activations of the network into the input units introduces *recurrence* into the network. This complicates the training process, because the output of the network is not correct, when the training starts and therefore, it cannot be fed back directly, when the training starts. Instead a weighted average of the actual output and the target output is used. It resembles more the output of the trained network which is similar (or at least should be similar) to the target output. At the beginning of the training, the weighting of the target output is high. It falls to zero during the training.

The network is trained on a tagged corpus. Target activations are 0 for all output units, excepted for the unit which corresponds to the correct tag, for which it is 1. A slightly modified version of the backpropagation algorithm with momentum term which has been presented in the last section is used: if the difference between the activation of an output unit $j$ and the corresponding target output is below a predefined threshold (we used 0.1), the error signal $\delta_{pj}$ is set to zero. In this way the network is forced to pay more attention to larger error signals. This resulted in an improvement of the tagging accuracy by more than 1 percent.

Network architectures with and without hidden layers have been trained and tested. In general, MLP-networks with hidden layers are more powerful than networks without one, but they also need more training and there is a higher risk of overlearning[4]. As will be shown in the next section, the Net-Tagger did not profit from a hidden layer.

In both network types, the tagging of a single word is performed by copying the tag probabilities of the current word and its neighbours into the input units, propagating the activations through the network to the output units and determining the output unit which has the highest activation. The tag corresponding to this unit is then attached to the current word.

If the second strongest activation in the output layer is close to the strongest one, the tag corresponding to the second strongest activation may be given as an alternative output. No additional computation is required for this. Further, it is possible to give a scored list of all tags as output.

---

[3]The output activations of the network do not necessarily sum to 1. Therefore, they should not be interpreted as probabilities.

[4]Overlearning means that irrelevant features of the training set are learned. As a result, the network is unable to generalize.

## 5 THE LEXICON

The lexicon which contains the a priori tag probabilities for each word is similar to the lexicon which was used by Cutting et al. (1992). It has three parts: a *fullform lexicon*, a *suffix lexicon* and a *default entry*. No documentation of the construction algorithm of the suffix lexicon in (Cutting et al., 1992) was available. Thus, a new method based on information theoretic principles was developed.

During the lookup of a word in the lexicon of the Net-Tagger, the fullform lexicon is searched first. If the word is found there, the corresponding tag probability vector is returned. Otherwise, the uppercase letters of the word are turned to lowercase, and the search in the fullform lexicon is repeated. If it fails again, the suffix lexicon is searched next. If none of the previous steps has been successfull, the default entry of the lexicon is returned.

The fullform lexicon was created from a tagged training corpus (some 2 million words of the Penn Treebank Corpus). First, the number of occurrences of each word/tag pair was counted. Afterwards, those tags of each word with an estimated probability of less than 1 percent were removed, because they were in most cases the result of tagging errors in the original corpus.

Figure 3: A sample suffix tree of length 3

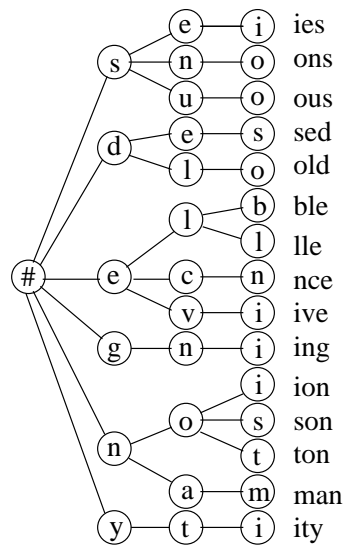

The second part of the lexicon, the suffix lexicon, forms a tree. Each node of the tree (excepted the root node) is labeled with a character. At the leaves, tag probability vectors are attached. During a lookup, the suffix tree is searched from the root. In each step, the branch which is labeled with the next character from the end of the word suffix, is followed.

Assume e.g., we want to look for the word *tagging* in the suffix lexicon which is shown in fig. 3. We start at the root (labeled #) and follow the branch which leads to the node labeled *g*. From there, we move to the node labeled *n*, and finally we end up in the node labeled $i$. This node is a leaf and the attached tag probability vector (which is not shown in fig. 3) is returned.

Table 1: Sample frequencies at a tree node and its two child nodes.

| tag | suffix *ess* | suffix *ness* | suffix *less* |
|---|---|---|---|
| JJ | 86 | 1 | 85 |
| NN | 10 | 2 | 8 |
| NP | 45 | 45 | 0 |
| RB | 2 | 0 | 2 |
| total | 143 | 48 | 95 |

The suffix lexicon was automatically built from the training corpus. First, a *suffix tree* was constructed from the suffices of length 5 of all words which were annotated with an open class part-of-speech[5]. Then tag frequencies were counted for all suffices and stored at the corresponding tree nodes.

In the next step, an *information measure* $I(S)$ was calculated for each node of the tree:

$$I(S) = -\sum_{pos} P(pos|S) \, log_2 P(pos|S) \quad (7)$$

Here, $S$ is the suffix which corresponds to the current node and $P(pos|S)$ is the probability of tag *pos* given a word with suffix $S$.

Using this information measure, the suffix tree has been pruned. For each leaf, the weighted *information gain* $G(aS)$ was calculated:

$$G(aS) = F(aS) \, (I(S) - I(aS)), \quad (8)$$

where $S$ is the suffix of the parent node, $aS$ is the suffix of the current node and $F(aS)$ is the frequency of suffix $aS$.

If the information gain at some leaf of the suffix tree is below a given threshold[6], it is removed. The tag frequencies of all deleted subnodes of a parent node are collected at the *default node* of the parent node. If the default node is the only remaining subnode, it is deleted too. In this case, the parent node becomes a leaf and is also checked, whether it is deletable.

To illustrate this process consider the following example, where *ess* is the suffix of the parent node, *less* is the suffix of one child node and *ness* is the suffix of the other child node. The tag frequencies of these nodes are given in table 1.

The information measure for the parent node is:

$$I(ess) = -\frac{86}{143}log_2\frac{86}{143} - \frac{10}{143}log_2\frac{10}{143} - ... \approx 1.32 \quad (9)$$

The corresponding values for the child nodes are 0.39 for *ness* and 0.56 for *less*. Now, we can determine the weighted information gain at each of the child nodes. We get:

$$G(ness) = 48(1.32 - 0.39) = 44.64 \quad (10)$$

---
[5]Open class parts-of-speech are those, which allow for the production of new words (e.g. noun, verb, adjective).
[6]We used a gain threshold of 10.

Table 2: Comparison of recognition rates

| method | accuracy |
|---|---|
| Net-Tagger | 96.22 % |
| trigram tagger | 96.06 % |
| HMM tagger | 94.24 % |

$$G(less) = 95(1.32 - 0.56) = 72.20 \quad (11)$$

Both values are well above a threshold of 10, and therefore none of them should be deleted.

As explained before, the suffix tree is walked during a lookup along the path, where the nodes are annotated with the letters of the word suffix in reversed order. If at some node on the path, no matching subnode can be found, and there is a default subnode, then the default node is followed. If a leaf is reached at the end of the path, the corresponding tag probability vector is returned. Otherwise, the search fails and the default entry is returned.

The *default entry* is constructed by subtracting the tag frequencies at all leaves of the pruned suffix tree from the tag frequencies of the root node and normalizing the resulting frequencies. Thereby, relative frequencies are obtained which sum to one.

## 6 RESULTS

The 2-layer version of the Net-Tagger was trained on a 2 million word subpart of the Penn-Treebank corpus. Its performance was tested on a 100,000 word subpart which was not part of the training corpus. The settings of the network parameters were as follows: the number of preceding words in the context $p$ was 3, the number of following words $f$ was 2 and the number of training cycles was 4 millions. The training of the tagger took one day on a Sparc10 workstation and the tagging of 100,000 words took 12 minutes on the same machine.

In table 2, the accuracy rate of the Net-Tagger is compared to that of a trigram based tagger (Kempe, 1993) and a Hidden Markov Model tagger (Cutting et al., 1992) which were trained and tested on the same data. In order to determine the influence of the size of the training sample, the taggers were also trained on corpora of different sizes and tested again[7]. The resulting percentages of correctly tagged words are shown in figure 4.

These experiments demonstrate that the performance of the Net-Tagger is comparable to that of the trigram tagger and better than that of the HMM tagger. They further show that the performance of the Net-Tagger is less affected by a small amount of training data than that of the trigram tagger. This may be due to a much smaller number of parameters in the Net-Tagger: while the trigram tagger must accurately

---
[7]For this test, a slightly simpler network structure with two preceding and one following word in the input context was used.

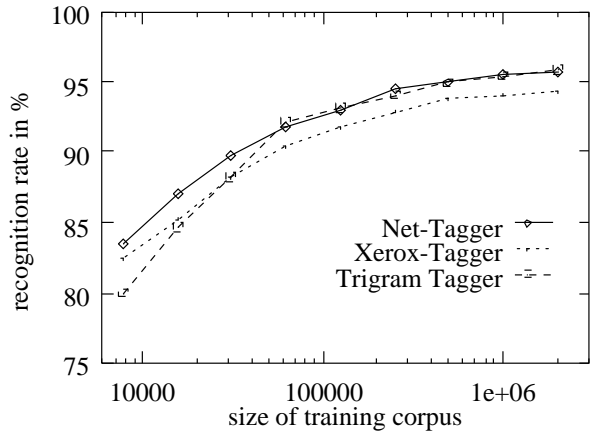

Figure 4: Recognition rates for varying sizes of the training corpus.

estimate 110,592 trigrams, the Net-Tagger only has to train 13,824 network parameters.

It was further tested, whether an additional hidden layer in the network with 50 units would improve the accuracy of the tagging. It turned out that the accuracy actually deteriorated slightly, although the number of training cycles had been increased to 50 millions[8].

Also, the influence of the size of the input context was determined. Shrinking the context from three preceding and two following words to two preceding and one following word reduced the accuracy only by 0.1 %. Enlarging the context gave no improvement. A context of three preceding and two following words seems to be optimal.

As mentioned previously, the tagger can produce an alternative tag, if the decision between two tags is difficult. In that way, the accuracy can be raised to 97.79 % at the expense of 4.6 % ambiguously tagged words.

An analysis of the errors of the Net-Tagger and the trigram tagger shows that both have problems with the same words, although the individual errors are often different[9].

## 7 CONCLUSIONS

In this paper, the Net-Tagger was presented, a part-of-speech tagger which is based on a MLP-network. A comparison of the tagging results with those of a trigram tagger and a HMM tagger showed that the accuracy is as high as that of the trigram tagger and the robustness on small training corpora is as good as that of the HMM tagger. Thus, the Net-Tagger combines advantages of both of these methods.

The Net-Tagger has the additional advantage that problematic decisions between tags are easy to detect,

---

[8] Due to the large training times needed to train the 3-layer-network, no further tests have been conducted.

[9] Less than 60 % of the tagging errors were made in common by both taggers.

so that in these cases an additional tag can be given in the output. In this way, the final decision can be delayed to a later processing stage, e.g. a parser.

A disadvantage of the presented method may be its lower processing speed compared to statistical methods. In the light of the high speed of present computer hardware, however, this does not seem to be a serious drawback.

## 8 REFERENCES

Church, K. W. (1985). A stochastic parts program and noun phrase parser for unrestricted text. *Proceedings of the Second Conference on Applied Natural Language Processing*, p. 136-143.

Cutting, D., J. Kupiec, J. Pedersen and P. Sibun (1992). A practical part-of-speech tagger. *Proceedings of the Third Conference on Applied Natural Laguage Processing*, Trento, Italy (ACL), pages 133-140, 1992. Also available as Xerox technical report SSL-92-01.

Federici, S. and V. Pirrelli (1993). Analogical modelling of text tagging. *unpublished report*, Istituto di Linguistica Computazionale, Pisa, Italy.

Greene, B. B and G. M. Rubin (1971). Automatic grammatical tagging of English. *technical report*, Department of Linguistics, Brown University, Providence, Rhode Island.

Jelinek, F. (1985). Markov Source modeling of text generation". In J.K. Skwirzinski Ed., *Impact of Processing Techniques on Communication*, Nijhoff, Dordrecht.

Kempe, A. (1993). A stochastic Tagger and an Analysis of Tagging Errors. *Internal paper*. Institute for Computational Linguistics, University of Stuttgart.

Lippmann, R. P. (1989). Review of Neural Networks for Speech Recognition. *Neural Computation, Vol. 1*, p. 1-38.

Nakamura, M., K. Maruyama, T. Kawabata and K. Shikano (1990). Neural network approach to word category prediction for English texts. In H. Karlgren Ed., *COLING-90*, Helsinki University, p. 213-218.

Rumelhart, D. E. and J. L. McClelland (1984). *Parallel Distributed Processing*. MIT-Press, Cambridge, MA.